\documentclass[letterpaper, 10pt, conference]{ieeeconf}      

\IEEEoverridecommandlockouts  
\usepackage{amsmath,amsfonts,amssymb}
\usepackage{algorithm}
\usepackage{float}
\newfloat{algorithm}{t}{lop}
\usepackage{xcolor}
\usepackage{algorithmic}
\usepackage{array}
\usepackage{textcomp}
\usepackage{stfloats}
\usepackage{url}
\usepackage{verbatim}
\usepackage{graphicx}
\usepackage{cite}
\usepackage{booktabs}
\usepackage{todonotes}
\usepackage{subcaption}

\definecolor{highlander_blue}{RGB}{0,61,165}
\definecolor{olive}{RGB}{128,128,0}

\makeatletter
\newcommand\fs@spaceruled{\def\@fs@cfont{\bfseries}\let\@fs@capt\floatc@ruled
  \def\@fs@pre{\vspace{0.5\baselineskip}\hrule height.8pt depth0pt \kern2pt}%
  \def\@fs@post{\kern2pt\hrule\relax}%
  \def\@fs@mid{\kern2pt\hrule\kern2pt}%
  \let\@fs@iftopcapt\iftrue}
\makeatother

\begin{document}

\title{SeGuE: Semantic Guided Exploration for Mobile Robots}

\author{Cody Simons, Aritra Samanta, Amit K. Roy-Chowdhury, Konstantinos Karydis
\thanks{The authors are with the Dept. of Electrical and Computer Eng. at the Univ. of California, Riverside, 900 University Avenue, Riverside, CA 92521, USA. 
Email: {\tt\footnotesize\{csimo005, asama004, amitrc, karydis\}@ucr.edu}. 
We gratefully acknowledge the support of NSF \# IIS-1901379, \# CNS-2312395 and ONR \# N00014-18-1-2252, \# N00014-19-1-2264. Any opinions, findings, and conclusions or recommendations expressed in this material are those of the authors and do not necessarily reflect the views of the funding agencies.
}}

\maketitle

\begin{abstract}
The rise of embodied AI applications has enabled robots to perform complex tasks which require a sophisticated understanding of their environment. To enable successful robot operation in such settings, maps must be constructed so that they include semantic information, in addition to geometric information. In this paper, we address the novel problem of semantic exploration, whereby a mobile robot must autonomously explore an environment to fully map both its structure and the semantic appearance of features. We develop a method based on next-best-view exploration, where potential poses are scored based on the semantic features visible from that pose. We explore two alternative methods for sampling potential views and demonstrate the effectiveness of our framework in both simulation and physical experiments. Automatic creation of high-quality semantic maps can enable robots to better understand and interact with their environments and enable future embodied AI applications to be more easily deployed.
\end{abstract}    
\section{Introduction}
%
%
Advancements in machine learning performance and edge computing propel robots toward acquiring higher than ever before reasoning and decision-making capabilities.  
These capabilities are critical for successfully deploying them in workplace environments such as automated kitchens~\cite{blodow2011kitchen} or unmanned farms~\cite{dechemi2023robotic}, where a robot needs to solve a joint navigation and semantic scene understanding problem. 

To solve this problem, classical metric-based maps have been fused with semantic-based maps to offer appropriate information-rich maps. 
For example, in visual object navigation~\cite{gadre2023cows}, an agent must find an object matching a text or picture. 
Similarly, in language-guided navigation~\cite{huang2023vlmaps}, an agent must navigate based on natural language instructions referencing the environment's appearance. 
Such methods are often zero-shot, assuming no prior knowledge of the environment. 
While this setting is challenging, it is inherently transient. 
In real-world cases, where continuous operation is needed, constructing a comprehensive map during or before operation would be highly beneficial. 
At the same time, several existing methods (e.g.,~\cite{blodow2011kitchen,achat2023semantic,gadre2023cows,huang2023vlmaps}) use maps that have been augmented with semantic classes. 
While semantic class information is valuable, its limited reusability may hinder transfer to physical robot deployment across environments. 
To address this challenge, we propose an exploration method that maps the environment's semantic \emph{features}, thus enabling reusability across diverse tasks. 

Modern AI-enhanced semantic maps can jointly encode a scene's geometry and appearance features. 
They can hold features extracted by machine learning models, which in turn can be reused across many different tasks. 
Semantic feature maps can be created online~\cite{gadre2023cows}, typically to aid in visual object search, or offline~\cite{huang2023vlmaps}, which can be used for language-guided navigation or other downstream tasks. 
However, offline mapping typically requires manual robot operation or hand-engineered routes to ensure comprehensive coverage of appearance features. 
Although autonomous exploration methods for mapping environmental structures are well-established in robotics, there is a notable gap in \emph{automatically guiding exploration to create high-quality semantic maps}. 

%
%
Autonomous environment exploration and mapping have received significant attention over the years. 
Mapping the structure of a static environment in 2D has largely been solved by existing SLAM approaches~\cite{cadena2016slam_survey}. 
Exploration is also very well studied, with conventional approaches divided broadly into frontier-based and next-best-view-based methods. 
In frontier-based methods~\cite{yamauchi1997frontiers}, the robot moves to locations that are on the border between the mapped and unmapped portions of the environment, while next-best-view methods~\cite{bircher2016rhnbv,palazzolo2018mavs,selin2019efficient}  sample and score potential views based on information theoretic metrics. Given sufficient time, both the methods can fully capture the geometry of an environment. However, they do not consider the appearance features of the environment. Furthermore, the increased range and field-of-view of modern depth sensors can result in incomplete and low-quality semantic feature maps \cite{brossard2020new}.

%
%
In this work, we introduce a novel autonomous exploration method based on the next-best-view approach, which maps the structure of the environment and simultaneously captures its semantic appearance. 
Our method scores potential views by computing the average entropy of visible semantic features, excluding occluded and converged features. 
We investigate two alternative view sampling techniques: uniform sampling across reachable poses and an iterative importance sampling method inspired by particle filter optimization. 
Off-the-shelf components are leveraged for metric-based mapping and navigation. 
Our approach is validated through both simulation and physical hardware experiments using a wheeled mobile robot.
In sum, our contributions include:
\begin{enumerate}
    \item We propose a method for scoring individual semantic map features and show how these scores can be combined to score a particular view.
    \item We explore two alternative methods for sampling potential views.
    \item We demonstrate the feasibility of our framework in simulated and physical experiments.
\end{enumerate}
\section{Related Work}
\subsection{Semantic  Mapping}
Semantic mapping has been explored both in machine learning and robotics communities. 
Machine learning research focuses on dense semantic features~\cite{huang2023vlmaps}. 
These maps allow for fine-grained search within the environment~\cite{huang2023vlmaps} and can be used to predict the layout of the unexplored portion of the environment~\cite{yokoyama2024vlfm}. 
Within robotics, there has been a focus on semantic classes. 
In~\cite{chen2019suma++}, semantic categories were used to increase loop closures, and in~\cite{liu2023heuristic_sampling}, semantic categories were used to create more efficient routes. 
As resources available on robots (sensors, computing units, etc.) continue to grow and AI models continue to become more amenable to edge computing, the demand for maps of dense semantic features is expected to rise. 
Our method can help construct such maps automatically.

\subsection{Frontier Exploration}
Frontier-based exploration has been thoroughly explored in the literature. 
Initially proposed in~\cite{yamauchi1997frontiers}, frontier-based methods seek to fully explore an environment by constantly moving toward the border between known and unknown space. 
More recent works have sought to increase the computational efficiency of computing frontiers~\cite{keidar2014efficient_frontiers} and create frontier ranking mechanisms~\cite{lubanco2020novel_frontiers}. 
These exploration methods are solely concerned with mapping the geometry of an environment, which can be directly observed with current LiDAR sensors that are commonly integrated into contemporary robots. 
In contrast, mapping semantic features can be significantly complicated by the variability in object appearances at different distances, the salience of various objects to feature extractors, and the limited field of view of most cameras. 

\subsection{Next-Best-View Determination}
The next-best-view pipeline~\cite{bircher2016rhnbv,palazzolo2018mavs,selin2019efficient} consists of a pose sampler, a metric to rank poses, either in terms of information gain~\cite{bircher2016rhnbv,palazzolo2018mavs,selin2019efficient} or a task-specific metric~\cite{achat2023semantic,blodow2011kitchen}, and a mapper which updates an internal map while navigating to the determined next-best-view pose. 
Several contemporary implementations use sampling-based planners to select poses~\cite{bircher2016rhnbv,steinbrink2021rrg}. 
However, this can lead to getting trapped in local maxima. 
One way to address this is by integrating an additional global sampler~\cite{selin2019efficient}. 
Information-theoretic metrics~\cite{monica2017contour,palazzolo2018mavs} encode the structure and the confidence in the current map to rank the value of a particular view. 
Semantic categories have also been used~\cite{blodow2011kitchen} to focus on more task-relevant structures. 
To the best of our knowledge, no method has incorporated a semantic feature map into a next-best-view algorithm.
\section{Problem Definition}
In this work, we address a novel semantic exploration problem. We aim to simultaneously explore and map an environment in real time autonomously while ensuring that the map fully captures the semantic information necessary for downstream embodied AI applications. 
Specifically, we ask how a mobile robot, equipped with an RGB camera and a depth sensor, can explore an environment and construct a 2D map $\mathcal{M}\in\mathbb{R}^{H\times W\times N+1}$, where $H$ and $W$ represent the height and width of the map, and $N$ represents the dimension of the semantic feature vectors. 
The additional scalar value in the third dimension represents the occupancy information, in the form of an occupancy grid. 
While semantic map generation is a thoroughly studied field~\cite{huang2023vlmaps}, in this work, we specifically address the exploration problem to ensure sufficient coverage of the environment and to capture high-quality semantic features. 
To reduce implementation complexity, we make the working assumptions that the robot operates in a physically bounded environment, and that the environment is relatively flat and traversable. 
These assumptions are not restrictive since they relate to most common indoor navigation paradigms (as we test herein), and can directly carry over to outdoor navigation in structured environments. 

\section{Proposed Approach}\label{method}
We propose the Semantic Guided Exploration (SeGuE) method for mobile robots. 
Our method builds upon next-best-view algorithms. 
During exploration, a semantic map, $S$, is constructed using data from two different sensing modalities, specifically an RGB camera and a 3D LiDAR. 
SeGuE samples collision-free and reachable poses from the occupancy grid generated during navigation, based on the current semantic understanding of the environment. 
To estimate the potential information gain from each pose, the average prediction entropy of each semantic features visible from a pose is computed. 
To ensure that cells with higher aleatoric uncertainty are not overly exploited, the entropy of each semantic feature is tracked over time and features whose entropy has converged are excluded from score calculations. 
This process is repeated until a pose with a score above some threshold cannot be found. 
SeGuE is summarized in Algorithm \ref{algorithm}. 
We elaborate on each step of the approach next.


\floatstyle{spaceruled}
\restylefloat{algorithm}
\begin{algorithm}
    \caption{SeGuE Algorithm}\label{algorithm}
    \begin{algorithmic}[1]
        \REQUIRE Termination Threshold $\tau$
        \STATE Initialize an empty semantic map $S$ and empty occupancy map $O$
        \STATE Get image $img$, point cloud $ptCloud$, and current pose $p_{curr}$ from robot
        \STATE $S.\text{update}(img, ptCloud, p_{curr})$
        \STATE $O.\text{update}(ptCloud, p_{curr})$
        \WHILE{True}
            \STATE Sample a set of pose $P=\{p_i\}_{i=0}^N$ using $\text{PoseSampler}(S, O)$
            \STATE $p^*=\max_{p\in P} \text{PoseScore}(p, S, O)$
            
            \IF{$\text{PoseScore}(p^*, S, O)<\tau$}
                \STATE End exploration
            \ENDIF

            \STATE Begin navigating to pose $p^*$
            \WHILE{$p^*$ not reached}
                \STATE Get image $img$, point cloud $ptCloud$, and current pose $p_{curr}$ from robot
                \STATE $S.\text{update}(img, ptCloud, p_{curr})$
                \STATE $O.\text{update}(ptCloud, p_{curr})$
            \ENDWHILE
        \ENDWHILE
    \end{algorithmic}
\end{algorithm}

\subsection{Pose Scoring}
\label{method:view_scoring}
Poses are scored according to the amount of information that can be gained at that pose. 
We use entropy as a proxy for this, encouraging the robot to explore the areas of the map it is most uncertain about. During the exploration process, the semantic score map is continually updated by extracting dense semantic features from the current image observation. These features are then associated with different points in the point cloud \textit{ptCloud}. 
As the 3D points now represent semantic features, they are projected onto the 2D semantic map to associate each cell with a feature. 
(Details of this procedure can be  found in~\cite{huang2023vlmaps}.) 
The semantic features are extracted from the image \textit{img} using a machine learning model $h=f\circ g$, which can be decomposed into a feature extractor $f$ and classifier $g$. 
Let $g$ be able to classify $M$ classes. 
The features stored in our semantic map are extracted from $f$. 
Throughout our experiments we make use of DinoV2~\cite{oquab2023dinov2} to extract features and a linear classifier fine-tuned on the ADE20K dataset~\cite{zhou2017ade}. 
The rest of the section describes the scoring mechanism in detail.

\subsubsection{Feature Scoring}\label{method:semantic_score}
Each cell in the semantic map contains a semantic feature vector that encodes the appearance of that location. 
The quality of every semantic feature associated with a particular cell changes over time, as the corresponding viewpoint varies. 
The prediction entropy of each cell is computed to assess the quality of the semantic feature. 
Using the classification head $g$, which classifies $M$ classes, the score for a semantic feature $x$ is
\begin{equation}\label{eq:score}
    s(x) = \frac{H(g(x))}{H(\mathcal{U}(M))}\;,
\end{equation}
where $H$ is the Shannon entropy and $\mathcal{U}$ is the uniform distribution. 
The score is normalized by the maximum entropy to be in the range $[0,1]$, since the maximum entropy can grow arbitrarily large as $M$ increases. 
Any cell with no semantic feature is assigned a score of $1$, indicating that no observations in a cell make it maximally uncertain. 
Once these scores are computed for every cell in the semantic map, potential views can be ranked by averaging the scores of individual cells that are visible from that viewpoint.

\subsubsection{Visibility Mask}\label{method:visibility_mask}
\begin{figure}[!t]
\captionsetup{font=footnotesize}
\vspace{6pt}
\centering
\includegraphics[width=0.6\columnwidth]{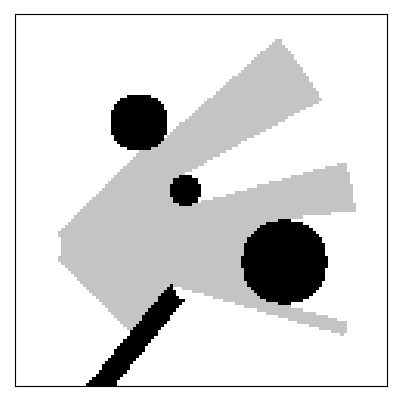}
\vspace{-6pt}
\caption{An example of the view mask generated by our raytracing algorithm. Known obstacles are shown in black and the view mask is shown in gray.}
\label{fig:view_mask}
\vspace{-12pt}
\end{figure}

We require that only locations that are visible from a given pose should be considered when providing a score for that pose. 
To enforce this, a raytracing algorithm, similar to~\cite{yokoyama2024vlfm}, is used to construct a view-mask of cells visible from a particular pose. 
An example mask is shown in Fig.~\ref{fig:view_mask}. 
In the figure, only the cells highlighted in gray are included in the aforementioned averaging operation; all other cells are ignored. 
This ensures that the overall score of a pose is not affected by the individual subscores that may be gathered from cells that provide no new information.


\subsubsection{Feature Convergence Tracking}\label{method:convergence_tracking}
Even after having collected several observations, some cells may still have a high entropy due to the aleatoric uncertainty present in some classes. 
As such, in cases that additional observations would not improve a semantic feature score, then any new observation should not contribute to the pose score computation. 
Doing so can help prevent the exploitation of inherently uncertain cells. 
To address this, each cell's feature score is tracked over time and monitored for convergence.

A ratio convergence test is employed to decide when a feature score has converged. 
During exploration, when the feature score value of a non-converged cell is updated, the ratio $\frac{s_0}{s_1}$ is computed, where $s_0$ is the previous score and $s_1$ is the new score.
If the ratio is less than a user-defined threshold (set herein to $1.1$), then the score for that cell is considered to have converged. 
Converged features can still be updated if that feature is incidentally observed during exploration, but once the feature score has converged, the respective cell will no longer be used in pose scoring.

\subsection{Pose Sampling}\label{method:view_sampling}
To find the poses which maximize the scoring metric~\eqref{eq:score} and ensure the semantic information is fully mapped, diverse poses across the map must be considered. 
To that end, two pose sampling strategies, Uniform Sampling (US) and Importance Sampling (IS), are proposed. 
Both strategies employ the same scoring mechanism. 
Regardless of the sampling method, exploration is terminated if no pose is found within a certain percentage of the maximum possible score. 
This is also a user-defined hyperparameter of our method. 
The threshold is set to $5\%$ throughout our experiments.

\subsubsection{Uniform Sampling}\label{method:uniform_sampling}
In the first method, poses are sampled uniformly across the entire map, and unreachable poses are discarded. 
Sampling continues until a desired number of reachable poses is obtained. 
These poses are then scored according to~\eqref{eq:score}. 
The pose with the highest score is selected as the next pose to navigate to.

\subsubsection{Importance Sampling}\label{method:imporance_sampling}
While the US method can acquire poses with high entropy, a large number of samples need to be considered for assessment, which may be inefficient in practice. 
Inspired by~\cite{liu2016pfo}, we also implement a sampling approach based on particle filter optimization. 
This method samples an initial set of poses uniformly across reachable poses, which are used to fit a mixture of Gaussian models where the likelihood of each pose is directly proportional to its score~\eqref{eq:score}. 
This distribution is then used to sample a new set of poses, subject to the same reachability constraint, and the process repeats for a fixed number of iterations. 
The mean of the mixture component with the highest score is taken as the next pose to navigate to.

\section{Experiments}
We evaluate SeGuE in both simulated and real-world environments using a wheeled mobile robot. 
The entire software suite is written in the Robot Operating System (ROS) framework. 
Mapping and odometry estimation for localization are conducted using 3D LiDAR information via~\cite{teng2025adaptive}. 
Point cloud information is also projected onto the ground plane to obtain a 2D occupancy grid for motion planning. 
Waypoint navigation is conducted with the ROS built-in package \textit{move\_base}. 
The scene features are extracted via an RGB camera. 

To evaluate the quality of the semantic maps generated, we assess the map coverage and average entropy. 
Map coverage provides an insight into the proportion of the appearance features that have been captured during exploration. 
It is computed as the percentage of occupancy map cells, which are either free or occupied, that have a corresponding semantic feature. 
Higher values of map coverage are better. 
The average entropy indicates the quality of the observations across the map. 
Average entropy calculates the prediction entropy of the entire semantic feature map. 
Any cell observed in the occupancy map with no corresponding semantic feature is assumed to have maximum entropy. 
Lower values of average entropy are better.

\subsection{Simulation Results}
We simulate the Clearpath Jackal in Gazebo\cite{koenig2004gazebo} and the relevant sensors using the calibration data available in~\cite{teng2023centroid}. 
All simulation is carried out on a desktop computer equipped with an Intel i7 processor, an NVIDIA RTX 3090 and 32 GB of RAM. 
We show results in two different environments. 
The Small House environment replicates a residential home and the Bookstore environment is a retail space. 
For each environment, the initial conditions are kept fixed across SeGuE and each baseline method.

We compare SeGuE, using both the US and IS variants, against a representative frontier-based exploration method~\cite{Hörner2016} and a representative next-best-view method (RRG-NBV)~\cite{steinbrink2021rrg}. 
We also compare SeGuE to a \textit{No Score} baseline, where we replace our map scoring metric with a simple metric where every unseen cell has a score of one and every seen cell has a score of zero. 
The view scoring and importance sampling are kept fixed. 

The map quality metrics for each method are reported in Table~\ref{tab:simulation_results}.
It can be observed that, across the board, the frontier-based and RRG-NBV approaches do not perform well. 
This can be explained based on the fact that no appearance features are considered but instead scene geometry is mapped. 
Both SeGuE and the \textit{No Score} baselines are significantly better with lower average entropy and higher coverage of the semantic features, demonstrating the necessity of exploration methods that are aware of scene semantics. 
Our method of scoring semantic features helps lower the average entropy when combined with either sampling method. 

\begin{table}[!t]
\captionsetup{font=footnotesize}
\vspace{6pt}
    \centering
    \caption{Simulation Results.}
    \vspace{-3pt}
    \renewcommand{\arraystretch}{1.5}
    \resizebox{\columnwidth}{!}{
        \begin{tabular}{ccccccc}
            \toprule
            & \multicolumn{2}{c}{Small House} & \multicolumn{2}{c}{Bookstore}\\
            \cmidrule(lr){2-3}\cmidrule(lr){4-5}
            & Coverage $\uparrow$ & Average Entropy $\downarrow$ & Coverage $\uparrow$ & Average Entropy $\downarrow$ \\
            \midrule
            Frontiers Based~\cite{Hörner2016} & $0.303$ & $3.619$ & $0.477$ & $2.862$\\
            RRG-NBV~\cite{steinbrink2021rrg} & $0.730$ & $1.643$ & $0.255$ & $3.917$\\
            No Score, US & $0.905$ & $0.806$ & $0.925$ & $0.801$\\
            No Score, IS & $0.897$ & $0.842$ & $0.929$ & $0.787$\\
            SeGuE, US & $0.938$ & $0.673$ & $0.940$ & $0.723$\\
            SeGuE, IS & $0.932$ & $0.702$ & $0.939$ & $0.730$\\
            \bottomrule
        \end{tabular}
    }
    \label{tab:simulation_results}
    \vspace{-12pt}
\end{table}

\begin{figure*}[!t]
\captionsetup{font=footnotesize}
\vspace{6pt}
    \centering
    \subfloat[Frontier-based Occupancy]{\includegraphics[width=0.24\textwidth,trim={10 10 7 10},clip]{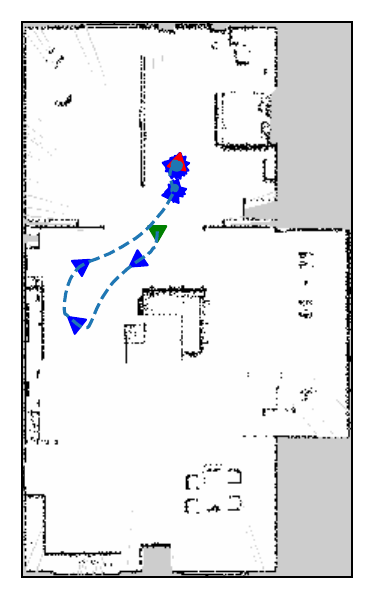}}%
    \hspace{0.005\textwidth}
    \subfloat[Frontier-based Semantics]{\includegraphics[width=0.24\textwidth,trim={79 28 79 48},clip]{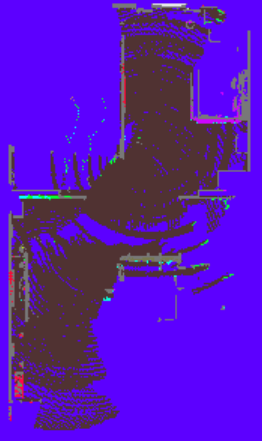}}%
    \hspace{0.005\textwidth}
    \subfloat[SeGuE Occupancy]{\includegraphics[width=0.24\textwidth,trim={10 10 7 10},clip]{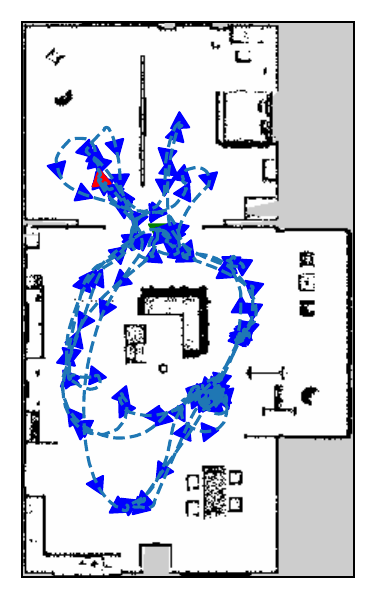}}%
    \hspace{0.005\textwidth}
    \subfloat[SeGuE Semantics]{\includegraphics[width=0.24\textwidth,trim={65 40 41 40},clip]{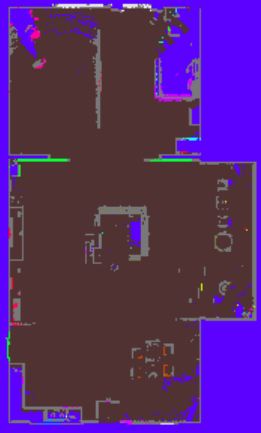}}%
    \caption{Visualization of the mapping results of (a-b) the frontier-based baseline and (c-d) SeGuE in the Small House environment. We show both the occupancy grid and semantic prediction, computed from the map of semantic features. On the occupancy grid, we plot the trajectory of the robot.}
    \label{fig:small_house_map_visualization}
\end{figure*}

\begin{figure*}[!t]
\captionsetup{font=footnotesize}
    \centering
    \subfloat[Frontier-based Occupancy]
{\includegraphics[width=0.24\textwidth,trim={10 9 7 10},clip]{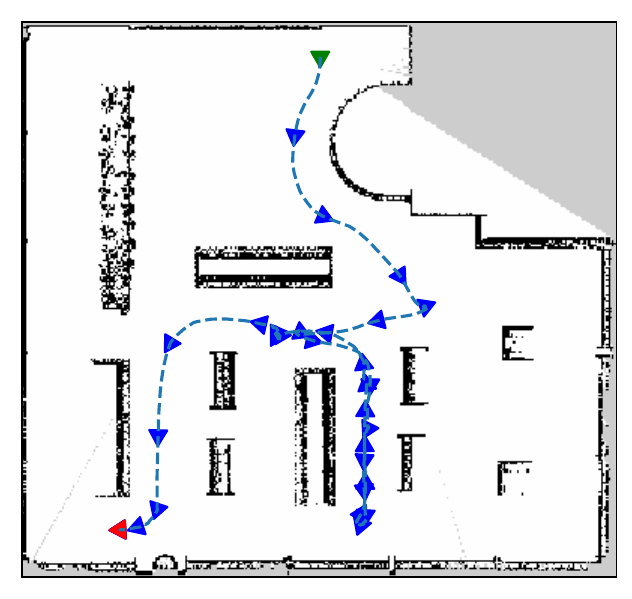}}%
    \hspace{0.005\textwidth}
    \subfloat[Frontier-based Semantics]
{\includegraphics[width=0.24\textwidth,trim={30 21 30 105},clip]{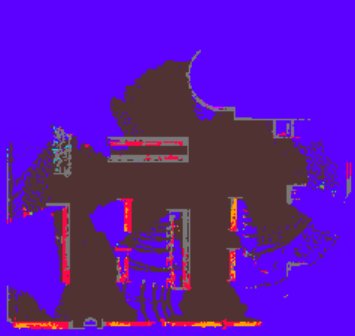}}%
    \hspace{0.005\textwidth}
    \subfloat[SeGuE Occupancy]    {\includegraphics[width=0.24\textwidth,trim={10 9 7 10},clip]{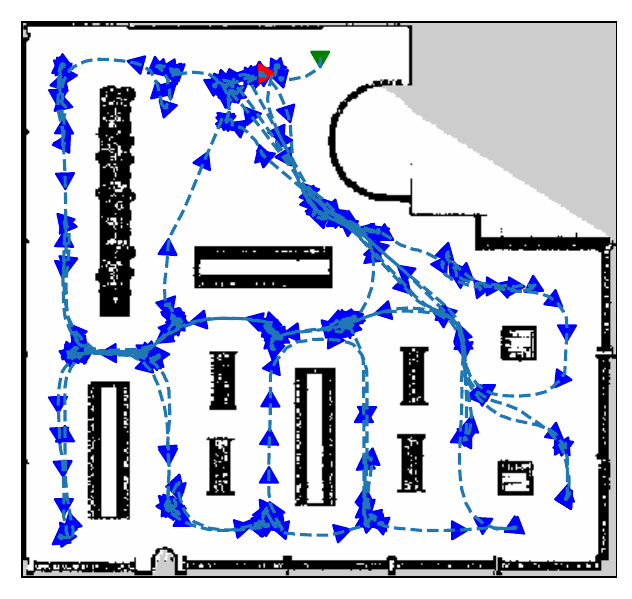}}%
    \hspace{0.005\textwidth}
    \subfloat[SeGuE Semantics]
{\includegraphics[width=0.24\textwidth,trim={30 21 30 105},clip]{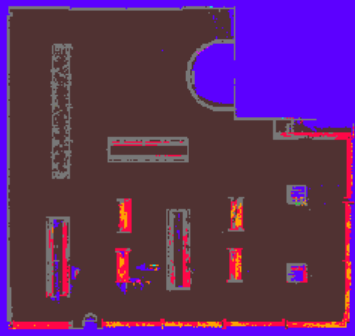}}%
    \caption{Visualization of the mapping results of (a-b) the frontier-based baseline and (c-d) SeGuE in the Bookstore environment. We show both the occupancy grid and semantic prediction, computed from the map of semantic features. On the occupancy grid, we plot the trajectory of the robot.}    \label{fig:bookstore_map_visualization}
    \vspace{-14pt}
\end{figure*}

We visualize the maps produced in each environment using the frontier-based method and SeGuE in Fig.~\ref{fig:small_house_map_visualization} (for Small House) and Fig.~\ref{fig:bookstore_map_visualization} (for Bookstore). 
Since the semantic features exist in a high-dimensional abstract feature space, we pass the features through a prediction head and visualize the class predictions, where each color represents a different class. 
It can be observed that the structure of the semantic feature map corresponds well with the structure of objects mapped in the occupancy grid. 

\subsection{Sample Number Study} 
We evaluate the sensitivity of both considered sampling methods to the number of samples chosen in the Small House environment. 
In the case of the uniform sampling method, this is a direct sweep over the number of samples chosen. 
The results of this ablation study, both with and without our semantic scoring method, are shown in Table~\ref{tab:sweep_us}. 
We note that, overall, our semantic scoring method increases coverage and decreases average entropy. 
While no direct relation between the number of samples and the performance appears to exist, we observe that, both with and without our semantic scoring method, the optimal performance seems to be between 100 and 200 samples. 
This is likely because too few samples do not approach an optimal score, while too many samples tend to over-exploit, leading to less exploration overall. 
\begin{table}[H]
\captionsetup{font=footnotesize}
    \centering
    \caption{Ablation Study: Sample Number in US.}
    \renewcommand{\arraystretch}{1.5}
    \resizebox{\columnwidth}{!}{
        \begin{tabular}{ccccc}
            \toprule
            & \multicolumn{2}{c}{Semantic Score} & \multicolumn{2}{c}{No Score}\\
            \cmidrule(lr){2-3}\cmidrule(lr){4-5}
            Samples & Coverage $\uparrow$ & Average Entropy $\downarrow$ & Coverage $\uparrow$ & Average Entropy $\downarrow$ \\
            \midrule
            10 & $0.897$ & $0.842$ & $0.895$ & $0.850$\\
            50 & $0.928$ & $0.711$ & $0.899$ & $0.837$\\
            100 & $0.908$ & $0.801$ & $0.911$ & $0.775$\\
            200 & $0.938$ & $0.673$ & $0.905$ & $0.806$\\
            500 & $0.904$ & $0.812$ & $0.887$ & $0.903$\\
            1000 & $0.912$ & $0.772$ & $0.895$ & $0.851$\\
            \bottomrule
        \end{tabular}
    }
    \label{tab:sweep_us}
\end{table}

We perform the same study on the IS method, varying the number of samples chosen every iteration and the total number of iterations. 
We report the quantitative results in Table~\ref{tab:sweep_is}.

When comparing the \textit{No Score} performance between sampling methods (cf. Tables~\ref{tab:sweep_us} and~\ref{tab:sweep_is}), the performance does not change significantly. 
We observe a marked improvement when the \textit{Semantic Score} is combined with IS, indicating that IS can better exploit the information within our metric. 
\begin{table}[H]
\captionsetup{font=footnotesize}
    \centering
    \caption{Ablation Study: Number of Samples and Iterations in IS.}
    \renewcommand{\arraystretch}{1.5}
    \resizebox{\columnwidth}{!}{
        \begin{tabular}{cccccc}
            \toprule
            & & \multicolumn{2}{c}{Semantic Score} & \multicolumn{2}{c}{No Score}\\
            \cmidrule(lr){3-4}\cmidrule(lr){5-6}
            Iterations & Samples & Coverage $\uparrow$ & Average Entropy $\downarrow$ & Coverage $\uparrow$ & Average Entropy $\downarrow$ \\
            \midrule
            2 & 50 & $0.932$ & $0.702$ & $0.897$ & $0.842$\\
            5 & 20 & $0.926$ & $0.704$ & $0.900$ & $0.844$\\
            10 & 10 & $0.905$ & $0.807$ & $0.881$ & $0.921$\\
            10 & 20 & $0.918$ & $0.750$ & $0.888$ & $0.876$\\
            10 & 50 & $0.919$ & $0.771$ & $0.916$ & $0.761$\\
            10 & 100 & $0.898$ & $0.836$ & $0.898$ & $0.850$\\
            \bottomrule
        \end{tabular}
    }
    \label{tab:sweep_is}
\end{table}

\subsection{Real-world Experiments}
In addition to the simulation studies, we evaluate SeGuE in a real-world environment (campus cafeteria). 
We use the Clearpath Jackal wheeled mobile robot equipped with the Velodyne VLP-16 LiDAR and Zed2i stereo camera; for a full description on sensor calibration see~\cite{teng2023multimodal}. 
Additionally, we connect the Clearpath Jackal robot with a laptop equipped with an Intel i7 processor, an NVIDIA RTX 3060, 32 GB of RAM and 512 GB of storage.

\begin{figure}[!t]
\vspace{6pt}
\captionsetup{font=footnotesize}
    \centering
    \subfloat[SeGuE Occupancy]{\includegraphics[width=0.22\textwidth,trim={155 11 161 45},clip]{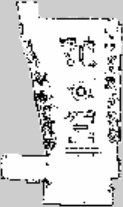}}%
    \hspace{0.01\textwidth}
    \subfloat[SeGuE Semantics]{\includegraphics[width=0.22\textwidth,trim={37 10 71 55},clip]{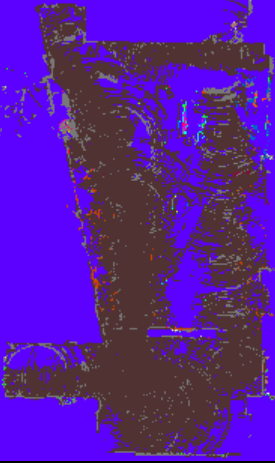}}%
    \caption{Mapping results of SeGuE in real-world experiments.}
    \label{fig:bytes_map_visualization}
    \vspace{-20pt}
\end{figure}

We tested SeGuE with uniform sampling since this was found to perform well consistently across simulation studies. 
Experimental results are shown in Fig.~\ref{fig:bytes_map_visualization}.
The computed coverage is $0.735$ and average entropy is $1.753$. 
While the overall coverage is lower compared to the values found in the simulation studies, our method can still map most of the environment. 
This performance degradation can be attributed to two factors. 
First, the underlying motion planning library did not transfer well to a more cluttered location. 
Second, battery constraints limited the total run time. 
Adjusting the underlying motion planner to be more energy-aware (e.g., by building on top of~\cite{kan2020online} and~\cite{kan2021task}) could enable optimal coverage subject to motion budget and task (i.e. viewpoint) constraints, so that the environment can be explored timely. 


\section{Conclusion}
We developed SeGuE, a method to autonomously explore an environment which ensures that appearance features are sufficiently observed so that a high-quality and complete map of semantic features may be constructed. 
We proposed a method to score the individual semantic features contained within the map and showed how these scores may be aggregated to rank potential viewpoints. 
Two viewpoint sampling methods were considered, with our tests showing how they interact with the system as a whole. 
Results in both simulation and physical hardware experiments showed that our method can fuse semantic feature information well, improving overall exploration metrics like map coverage and average entropy in comparison to baseline methods.

Since SeGuE builds on top of a next-best-view algorithm, it remains bound to inherent limitations of the latter, particularly the performance dependence on the sampling method and scoring. 
Further, we focus only on global sampling methods to avoid getting trapped in local minima. 
In future work, we aim to integrate local sampling as well.


\bibliographystyle{ieeetr}
\bibliography{main}

\end{document}